\documentclass[9pt,conference]{IEEEtran} %若不行需要修改上一行
\usepackage[margin=1.25in]{geometry}   
\usepackage{lmodern}
\usepackage{spconf,amsmath,graphicx,hyperref}
\usepackage{multirow}
\usepackage{booktabs}     
\usepackage{makecell}
\usepackage{cite}
\usepackage{placeins}
\usepackage{balance}
\usepackage{diagbox}
\title{Context-Aware Dynamic Chunking for Streaming Tibetan Speech Recognition}
\name{Chao Wang\textsuperscript{1,4}$^{\dagger}$,
      Yuqing Cai\textsuperscript{2,4}$^{\dagger}$,
       Renzeng Duojie\textsuperscript{3,4}$^{*}$, 
      Jin Zhang\textsuperscript{2,4}, 
      Yutong Liu\textsuperscript{2},
      Nyima Tashi\textsuperscript{1,2,3,4}$^{*}$}
\address{\textsuperscript{1} Qinghai Normal University, Xining, China \\
         \textsuperscript{2} University of Electronic Science and Technology of China, Chengdu, China \\
         \textsuperscript{3} Xizang University, Lhasa, China \\
         \textsuperscript{4} The State Key Laboratory of Tibetan Intelligence, Xining, China\\
        202433341016@stu.qhnu.edu.cn, \{202511090923, 202511090901\}@std.uestc.edu.cn, \\ yutongliu0620@foxmail.com, 
        \{rzdj, nmzx\}@utibet.edu.cn
        \thanks{$^{\dagger}$denotes equal contribution.}
        \thanks{$^{*}$Corresponding author.}}

\begin{document}
\ninept
\maketitle
\begin{abstract}
In this work, we propose a streaming speech recognition framework for Amdo Tibetan, built upon a hybrid CTC/Atten-tion architecture with a context-aware dynamic chunking mechanism. The proposed strategy adaptively adjusts chunk widths based on encoding states, enabling flexible receptive fields, cross-chunk information exchange, and robust adaptation to varying speaking rates, thereby alleviating the context truncation problem of fixed-chunk methods. To further capture the linguistic characteristics of Tibetan, we construct a lexicon grounded in its orthographic principles, providing linguistically motivated modeling units. During decoding, an external language model is integrated to enhance semantic consistency and improve recognition of long sentences. Experimental results show that the proposed framework achieves a word error rate (WER) of 6.23\% on the test set, yielding a 48.15\% relative improvement over the fixed-chunk baseline, while significantly reducing recognition latency and maintaining performance close to global decoding.
\end{abstract}

\section{Introduction}
\label{sec:intro}
\begin{figure}[t]
	\centering
	\includegraphics[width=0.9\linewidth]{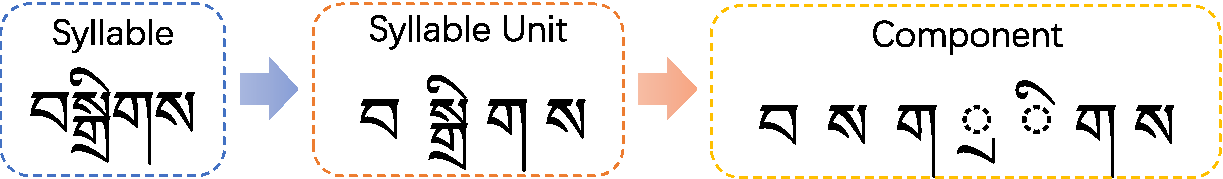}
	\caption{Structure of the Tibetan word \textit{Programming}.}
	\label{L1.eps}
\end{figure}

Automatic speech recognition (ASR) for morphologically complex and low-resource languages such as Tibetan\cite{qin2022finer,wu2024mpsa} remains challenging, primarily due to data scarcity, difficulties in selecting appropriate modeling units, and the demand for low-latency processing\cite{gong2025tibetan}. Tibetan is an alphasyllabary with a complex orthographic structure. A Tibetan character typically corresponds to a syllable, which is formed through the horizontal stacking of syllable unit, while each frame itself is vertically composed of smaller components. Accordingly, Tibetan modeling units can be defined at three levels—components, unit syllables, and syllables—as illustrated in Figure~\ref{L1.eps}. The Amdo dialect, in particular, is highly agglutinative and exhibits considerable variation in speaking rate, which makes accurate alignment difficult under conventional character- or phoneme-based approaches. Existing streaming ASR models largely rely on fixed-chunk mechanisms\cite{liu2023blockwise,zeyer2021monotonic,radford2023robust,moritz2021dual}, which struggle to capture long-range contextual dependencies and adapt to variable speech rates. Even advanced approaches such as MoChA\cite{xia2025mfla}, which introduce chunk-level attention, have not yet been systematically validated on Tibetan with distinct typological characteristics.

To address these challenges, we propose an end-to-end streaming recognition framework for Amdo Tibetan that integrates temporal and linguistic modeling. On the temporal side, we design a context-aware dynamic chunking mechanism that adaptively adjusts the attention window based on historical encoder states, enabling cross-chunk interaction and reducing boundary truncation errors. On the linguistic side, we construct three lexicons of different granularities grounded in Tibetan orthographic principles and conduct systematic comparisons against baseline models. Furthermore, we introduce both an external language model\cite{dighe2023leveraging} and a rescoring strategy\cite{shivakumar2025rescoring,chen2023largescale,udagawa2022effect} to improve recognition accuracy, particularly for long and syntactically complex sentences.

\begin{figure*}[htb]
    \centering
    \includegraphics[width=0.8\textwidth]{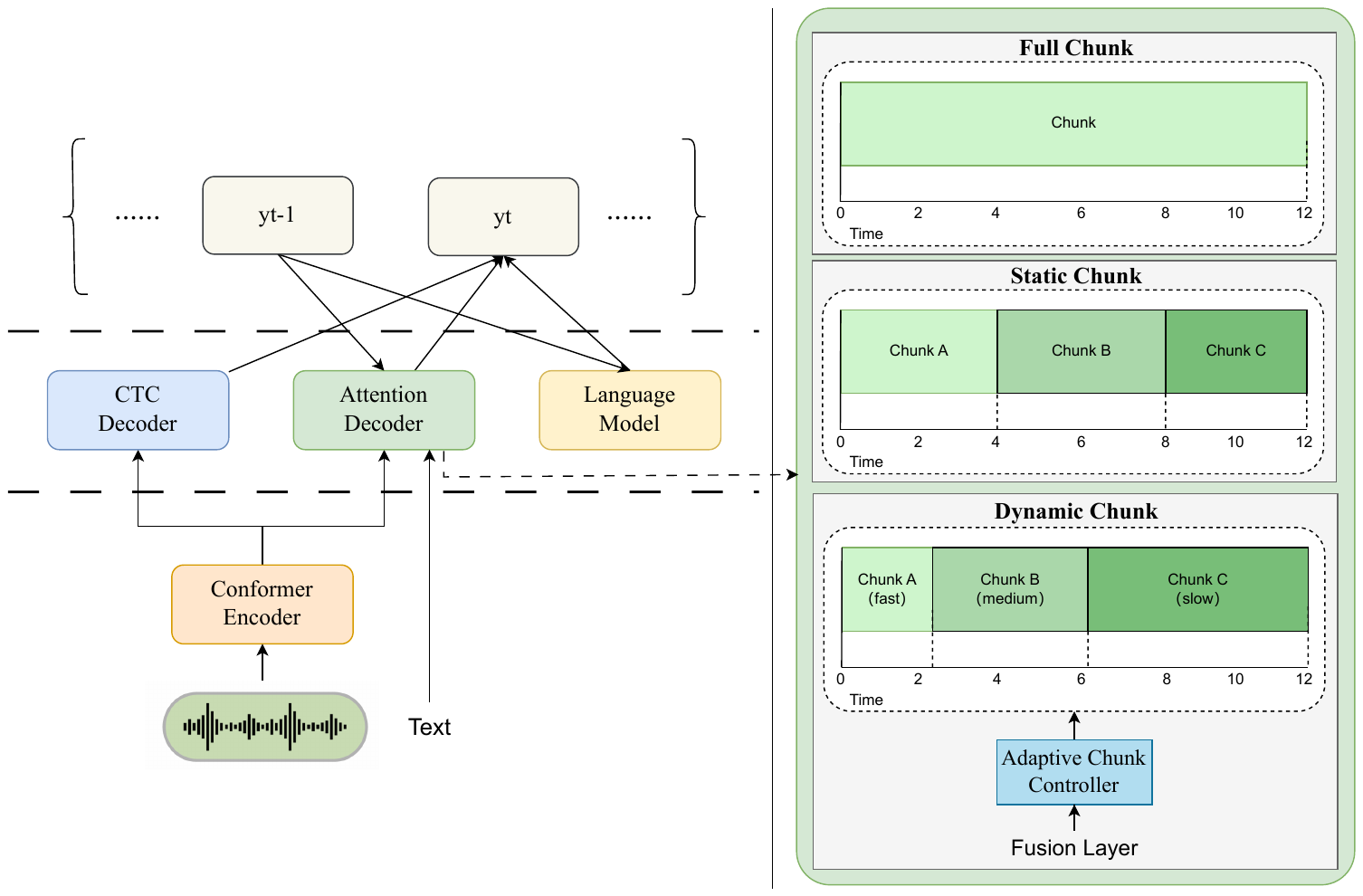}\vspace{-1.9cm}
    \caption{The overall model architecture is presented, along with a detailed illustration of the attention module specifically enhanced in this work to account for the phonetic characteristics of Tibetan.}
    \label{fig:res2}
\end{figure*}

 To facilitate future research, the lexicons, language model, and a portion of the Amdo Tibetan speech dataset developed in this work will be made publicly available at \url{https://github.com/chaonlplab/tibetanspeech}.

\section{METHODOLOGY}
\label{sec:format}
\subsection{Streamable Encoder-Decoder Architecture with Hybrid CTC/Attention}
As illustrated in Figure~\ref{fig:res2}, we adopt a hybrid CTC/Attention architecture\cite{gimeno2024comparison} to build an end-to-end recognition network tailored for streaming scenarios\cite{zeineldeen2023chunked}. The encoder is based on a Conformer structure and incorporates a cross-chunk contextual mechanism, which enables information transfer and dependency modeling across chunks\cite{dong2019self,shi2021emformer,chen2021streaming,swietojanski2023variable,gulzar2023ministreamer}. In conjunction with the proposed dynamic chunking strategy, the model adaptively adjusts its receptive field, thereby achieving strong contextual modeling while maintaining low latency. The decoder employs causal self-attention\cite{vaswani2017attention,brown2020language} to prevent access to future frames during streaming inference.

The training process jointly optimizes the CTC\cite{graves2006ctc} and Attention objectives:

\begin{equation}
L_{\text{total}} = \lambda L_{\text{CTC}} + (1 - \lambda) L_{\text{Attn}}
\end{equation}

where $\lambda$ denotes a weighting factor. Within this hybrid framework, the CTC branch provides global alignment constraints, while the Attention branch enhances fine-grained chunk-level modeling. In combination with dynamic chunking and cross-chunk context propagation, this design significantly improves the effectiveness of streaming recognition.

\subsection{Context-Aware Dynamic Chunking Mechanism and Training Method}
To achieve low-latency decoding in streaming speech recognition, 
we propose a context-aware dynamic chunking mechanism to replace the conventional fixed-window encoder input. 
During both training and inference, the chunk width is dynamically adjusted according to the contextual state, 
thereby balancing recognition accuracy and latency \cite{sheng2025dynamic}. 

At step $n$, the chunk width $W_n$ and stride $S_n$ are determined jointly by the encoding state of the previous chunk $h_{n-1}$ 
and the global context control vector $c_{n-1}$:

\begin{equation}
(W_n, S_n) = f_{\mathrm{ctrl}}(h_{n-1}, c_{n-1}),
\end{equation}

where $f_{\mathrm{ctrl}}(\cdot)$ is a learnable gating function implemented with a lightweight MLP \cite{rumelhart1986backprop} and an attention mechanism. Concretely, we first fuse local and global context:
\begin{align}
z_n &= \alpha_n \tanh(W_h h_{n-1}+W_c c_{n-1}) \nonumber\\
    &\quad + (1-\alpha_n)[h_{n-1};c_{n-1}],
\end{align}

with gating coefficient 
\[
\alpha_n = \sigma(w_\alpha^\top [h_{n-1};c_{n-1}]).
\]

Then $z_n$ is mapped into bounded continuous variables:
\begin{equation}
\hat{w}_n,\;\hat{s}_n = \sigma(W_2\phi(W_1 z_n+b_1)+b_2),
\end{equation}

which are finally discretized into valid integers:
\begin{align}
W_n &= \mathrm{round}\!\big(W_{\min}+(W_{\max}-W_{\min})\hat{w}_n\big), \\
S_n &= \mathrm{round}\!\big(S_{\min}+(S_{\max}-S_{\min})\hat{s}_n\big).
\end{align}

The controller automatically adjusts the chunk width according to speech rate and context complexity: 
it narrows the width to reduce latency when speech is fast and context is stable, and expands it to strengthen contextual modeling when speech is slow or the context is complex. 

Additionally, the chunk starting point can be aligned or slightly adjusted to improve decoding flexibility. 
Each chunk is processed independently by the encoder, 
and its output representation $h_n$ is passed to a higher-level attention module for cross-chunk dependency modeling.

To quantify latency, we adopt the Average Perceived Latency (APL)\cite{sharma2025unifying} metric:

\begin{equation}
\text{APL} = \frac{1}{L} \sum_{i=1}^{L} (t^\text{decode}_i - t^\text{input}_i),
\end{equation}

where $t^\text{decode}_i$ denotes the decoding completion time of the $i$-th label, and $t^\text{input}_i$ represents the earliest available input time for that label.

To fully exploit the advantages of multi-level modeling units and the proposed dynamic chunking mechanism, we adopt a three-stage training strategy. In the global training stage, we employ a Transformer with full attention to evaluate different modeling units—syllables, orthographic units, and sub-character components—in order to identify the most effective representation. Although this stage does not support streaming decoding, it enables optimal contextual modeling and provides a fair comparison across unit granularities.

\begin{table*}[htb]
\centering
\fontsize{9pt}{11pt}\selectfont % 设置字体为9pt
\caption{Comparison of global, fixed-chunk, and dynamic-chunk attention mechanisms for Tibetan speech recognition.(Att: Attention, Att-R: Attention-Rescoring, CTC-G: CTC-Greedy-Search, CTC-PBS: CTC-Prefix-Beam-Search).}
\label{tab:chunk_size}
\begin{tabular}{ccccccc|cccc}
\hline
\multirow{2}{*}{Chunk} & \multirow{2}{*}{Modeling Unit}  & \multirow{2}{*}{Size} & \multicolumn{4}{c|}{Test} & \multicolumn{4}{c}{Dev} \\
\cline{4-11}
 & & & Att & Att-Re & CTC-G & CTC-PBS & Att & Att-Re & CTC-G & CTC-PBS \\
\hline
\multirow{3}{*}{Global}& Syllable & \multirow{3}{*}{$\infty$} & \textbf{6.98} & - & - & - & \textbf{\underline{6.72}} & - & - & - \\
& Syllable Unit &   & 8.84 & - & - & - & 8.60 & - & - & - \\
& Component &   & 8.86 & - & - & - & 8.65 & - & - & - \\
\hline
\multirow{4}{*}{Static} 
& \multirow{4}{*}{Syllable} & 8  & - & 10.36 & 11.12 & 11.04 & - & 10.03 & 10.80 & 9.97 \\
& & 14 & - & 10.08 & 10.69 & 10.59 & - & 9.40 & 10.04 & 9.97 \\
& & 16 & - & \textbf{9.73} & \textbf{10.21} & \textbf{10.11} & - & \textbf{\underline{9.23}} & 9.99 & 9.96 \\
& & 20 & - & \textbf{9.73} & 10.28 & 10.21 & - & 9.27 & \textbf{9.83} & \textbf{9.77} \\
\hline
Dynamic& Syllable & - & - & \textbf{8.94} & 9.49 & 9.61 & - & \textbf{\underline{7.44}} & 9.23 & 9.29 \\
\hline
\end{tabular}
\end{table*}

After selecting the best-performing unit, the model proceeds to the subsequent stages. In the fixed-chunk training stage, a stable frame-to-label alignment is established using a fixed chunk width $W$ and stride $S$. The model is optimized with a joint CTC/Attention objective:

\begin{equation}
L_\text{static} = \lambda L_\text{CTC-static} + (1-\lambda)L_\text{Attn-static}.
\end{equation}

In the dynamic chunk training stage, the dynamic chunk controller $f_\text{ctrl}$ is activated, and a latency regularization term is incorporated into the objective function:

\begin{equation}
L_\text{dynamic} = L_\text{total} + \alpha \cdot \text{APL},
\end{equation}

where $\alpha$ is a weighting factor used to balance recognition accuracy and latency performance.

\section{EXPERIMENT}
\label{sec:pagestyle}
\subsection{Datasets}
A large-scale Amdo Tibetan speech corpus comprising approximately 1,000 hours was constructed, covering 2,000 speakers (roughly balanced gender, ages 18–60). The dataset construction pipeline was designed by following the procedure described in\cite{schultz2013globalphone} and recorded in diverse scenarios including news broadcasting, interviews, and Q\&A dialogues. Natural variations in speech rate, audio quality, and background noise are present. All utterances were transcribed, denoised, and sentence-aligned by Tibetan phonetics experts, and the training, validation, and test sets were randomly partitioned to ensure fair evaluation.The detailed statistics of the corpus are summarized in Table ~\ref{tab:dataset}.
\begin{table}[h]
\centering
\caption{Tibetan Amdo Dialect Corpus}
\label{tab:dataset}
{\fontsize{9pt}{10.8pt}\selectfont % 设置表格字体为9pt
\begin{tabular}{@{}c c c c@{}}
\toprule
 \diagbox[width=3.3em,height=1em]{}{} & \text{Train set} & \text{Dev set} & \text{Test set}\\
\midrule
Sentences & 1293538 & 100000 & 100000 \\
\bottomrule
\end{tabular}
} % 结束字体设置
\end{table}
\subsection{Setup}
Acoustic features consist of 80-dimensional Fbank\cite{davis1980fbank} extracted from audio sampled at 16 kHz, using a 25 ms frame length and a 10 ms frame shift. The recognition model consists of a 12-layer Conformer encoder coupled with a 6-layer bidirectional Transformer decoder. Each layer is configured with 1024 hidden units, 8 attention heads, a dropout rate of 0.1, and subsampling rates of 2/6/8. Model training is performed using the Adam optimizer with an initial learning rate of 0.002.

\subsection{Performance Analysis of Fixed and Dynamic Chunking}
We first employ a Transformer with full attention to compare three types of modeling syllable, syllable unit, and components—and select syllables as the primary modeling unit. Then evaluate the effect of fixed chunk sizes on recognition performance, as shown in Table~\ref{tab:chunk_size}. Among all settings, the rescoring strategy with a chunk size of 16 frames achieves the best result, yielding a word error rate (WER) of 9.23\%. This performance is close to that of full-context decoding and thus serves as an important reference baseline for the design of dynamic chunking.

After introducing dynamic chunking, the proposed approach consistently outperforms fixed chunking across all decoding strategies. Further analysis of historical context and the look-left mechanism (Table ~\ref{tab:Left-Side}) shows that when the cross-chunk context carry-over is set to 8, the validation WER reaches a minimum of 7.91\%, underscoring the critical role of left-context awareness in streaming recognition.

\begin{table}[th]
\centering
\caption{WER(\%) Comparison of Dynamic Blocking Left-Side Perception Mechanism}
\label{tab:Left-Side}
{\fontsize{9pt}{11pt}\selectfont
\begin{tabular}{cccccc}
\toprule
\multirow{2}{*}{Carry-over} & Decoding & \multicolumn{2}{c}{Dynamic-Look-left} & \multicolumn{2}{c}{Static} \\
\cmidrule(r){3-4} \cmidrule(l){5-6}
 & method & Test & Dev  & Test  & Dev  \\
\midrule
2 & \multirow{4}{*}{Att-Re} & 8.37 & 8.12 & 8.71 & 8.63 \\
4 &  & 8.25 & 7.98 & 8.58 & 8.51 \\
6 &  & 8.21 & 7.94 & 8.52 & 8.42 \\
8 &  & \textbf{8.18} & \textbf{\underline{7.91}} & \textbf{8.50} & \textbf{8.40} \\
\bottomrule
\end{tabular}
}
\end{table}

\subsection{Joint Optimization of CTC and Attention Weights}
The effect of the weighting coefficient $\lambda$ in the joint loss on recognition performance was examined, as shown in Table~\ref{tab:CTC_ratio}. The best performance is obtained at $\lambda$ = 0.5, yielding a WER of 7.28\% under the Attention-based decoding strategy. This setting achieves a balanced trade-off: the CTC branch ensures stable alignment, while the attention mechanism strengthens cross-chunk modeling. In contrast, excessively high or low values of $\lambda$ result in unstable training, thereby confirming the effectiveness of joint optimization.

% \begin{table}[th]
% \centering
% \caption{Experiment on CTC's Contribution in Joint Optimization(WER\%)}
% \label{tab:CTC_ratio}
% \resizebox{\columnwidth}{!}{%
% \begin{tabular}{cccccc}
% \toprule
% Model & $\lambda$& Attention & Attn-Rescoring & CTC-Greedy &CTC-Prefix \\
% \midrule
% \multirow{4}{*}{CCBT}&0.1 & 7.99 & 8.16 & 8.13 & 8.38 \\
%     &0.3 & 7.99 & 7.86 & 8.13 & 8.06 \\
%     &0.5 & 7.99 & 7.28 & 8.13 & 7.94 \\
%     &0.7 & 7.99 & 7.81 & 8.13 & 8.06 \\
% \bottomrule
% \end{tabular}%
% }
% \end{table}
\begin{table}[th]
\centering
\caption{Experiment on CTC's Contribution in Joint Optimization (WER\%)}
\label{tab:CTC_ratio}
{\fontsize{9pt}{11pt}\selectfont
\begin{tabular}{ccccc}
\toprule
$\lambda$ & Att & Att-Re & CTC-G & CTC-PBS \\
\midrule
0.1 & \multirow{4}{*}{\textbf{7.99}} & 8.16 & \multirow{4}{*}{\textbf{8.13}} & 8.38 \\
0.3 &  & 7.86 &  & 8.06 \\
0.5 &  & \textbf{\underline{7.28}} &  & \textbf{7.94} \\
0.7 &  & 7.81 &  & 8.06 \\
\bottomrule
\end{tabular}
}
\end{table}

\subsection{Latency Performance Evaluation}
Model latency was assessed using the Average Perceived Latency (APL) metric. The results show that larger chunk sizes or reduced inter-chunk overlap increase decoding latency. In contrast, the proposed dynamic chunking mechanism achieves strong recognition performance while keeping the latency as low as 0.78 seconds (with a 32-frame chunk size and 8-frame overlap), thereby meeting the requirements of real-time applications.

\begin{table}[h]
\centering
\caption{Statistics on syllable-level output latency based on the block-wise attention encoder-decoder model (Time unit: seconds)}
\label{tab:Latency}
{\fontsize{9pt}{10.8pt}\selectfont % 设置表格字体为9pt
\begin{tabular}{@{}c c c@{}}
\toprule
\text{Chunk Size} & \text{Overlap Frame Count} & \text{Delay Rate} \\
\midrule
8 & 0 & 1.92 \\
16 & 4 & 1.04 \\
32 & 8 & \textbf{\underline{0.78}} \\
\bottomrule
\end{tabular}
} % 结束字体设置
\end{table}

\subsection{Long-form Speech Generalization Capability}
To assess the model’s ability to generalize to long-form speech, multiple utterances were concatenated to create test sequences ranging from 1000 to 2000 frames. As shown in Table~\ref{tab:Long}, the chunk-based model maintains stable recognition accuracy, demonstrating that the proposed streaming architecture exhibits strong robustness in handling long-utterance recognition tasks.

\begin{table}[h]
\centering
\caption{WER(\%) in Long-Form Speech Recognition Research}
\label{tab:Long}
{\fontsize{9pt}{10.8pt}\selectfont % 设置表格字体为9pt
\begin{tabular}{@{}c c c@{}}
\toprule
\text{Frame rate} & \text{Test} & \text{Dev} \\
\midrule
1000 & \textbf{7.62} & \textbf{\underline{7.35}} \\
1500 & 7.71 & 7.48 \\
2000 &7.80 & 7.61 \\
\bottomrule
\end{tabular}
} % 结束字体
\end{table}

\subsection{Batch Size and Global Normalization Strategy}
We further evaluated the impact of different beam widths\cite{graves2012sequencetransduction} and the application of global normalization on recognition performance ( ~\ref{tab:beam_comparison}). The results show that the chunk-based model is particularly sensitive to beam width. Specifically, with a beam size of 10, applying global normalization\cite{lafferty2001crf} yields the lowest WER of 7.28\%, representing a substantial improvement over the non-normalized setting.

\begin{table}[thb]
\centering
\caption{Performance Comparison with Different Beam Sizes and Global Normalization}
\label{tab:beam_comparison}
\resizebox{\columnwidth}{!}{%
\begin{tabular}{cccccc}
\toprule
\text{Global Norm.} & \text{Beam Size} & \text{Att} & \text{Att-Re} & \text{CTC-G} & \text{CTC-PBS} \\
\midrule
\multirow{4}{*}{Yes} 
 & 5 & \multirow{4}{*}{\textbf{7.34}} & 8.05 & \multirow{4}{*}{\textbf{8.63}} & 8.61 \\
 & 10 &  & \textbf{\underline{7.28}} &  & \textbf{7.94} \\
 & 15 &  & 7.94 &  & 8.61 \\
 & 20 &  & 7.64 &  & 8.26 \\
\midrule
\multirow{4}{*}{No} 
 & 5 & \multirow{4}{*}{\textbf{7.34}} & 8.11 & \multirow{4}{*}{\textbf{8.63}} & 8.63 \\
 & 10 &  & 8.03 &  & 8.59 \\
 & 15 &  & 7.92 &  & 8.56 \\
 & 20 &  & \textbf{7.85} &  & \textbf{8.50} \\
\bottomrule
\end{tabular}%
}
\end{table}

\subsection{Language Model Fusion Effect}
We adopt a shallow fusion approach to integrate an n-gram language model, with the Part LM trained on 1.5 GB of Tibetan text and the Full LM on 3 GB (Table~\ref{tab:lm_comparison}). Results indicate that the Full LM reduces the WER of the dynamic chunk model from 7.28\% to 6.23\% and that of the full-context model from 6.98\% to 5.03\%. Furthermore, the Full LM consistently outperforms the Part LM, underscoring the importance of semantic consistency in Tibetan speech recognition and confirming that the proposed framework can approach full-context decoding performance under streaming conditions.
\begin{table}[h]
\centering
\caption{WER(\%) Comparison of Different Language Models in AED Systems}
\label{tab:lm_comparison}
{\fontsize{9pt}{10.8pt}\selectfont % 设置表格字体为9pt
\setlength{\tabcolsep}{8pt} % 调整列间距
\begin{tabular}{ccc}
\toprule
 \text{System Type} & \text{Language Model} & \text{WER} \\
\midrule
\multirow{3}{*}{Dynamic Chunk AED} 
 & - & 7.28 \\
 & Part\_lm & 6.41 \\
 & Full\_lm & \textbf{6.23} \\
\midrule
\multirow{3}{*}{Global AED} 
 & - & 6.98 \\
 & Part\_lm & 5.19 \\
 & Full\_lm & \textbf{\underline{5.03}} \\
\bottomrule
\end{tabular}
} % 结束字体设置
\end{table}

\section{CONCLUSION}
\label{sec:typestyle}

This paper presents an end-to-end streaming speech recognition framework for Amdo Tibetan, combining a context-aware dynamic chunking mechanism with linguistically motivated multi-level modeling units. Through systematic evaluation, syllables are identified as the most effective representation, and together with external language modeling and rescoring, the framework achieves performance comparable to full-context decoding while maintaining low latency. Supported by a newly constructed large-scale Amdo Tibetan corpus, the proposed approach provides a practical solution for streaming ASR in morphologically complex, low-resource languages, and the released lexicons, pretrained language model, and corpus subsets will further facilitate future research.

\section{ACKNOWLEDGEMENT}
 This work is supported by the National Science and Technology Major Project (No. 2022ZD0116100), the National Natural Science Foundation of China Key Program (No. 62436006), the National Natural Science Foundation of China Youth Program (No. 62406257), and the Natural Science Foundation of Tibet Autonomous Region (No. XZ202401ZR0031).

\FloatBarrier
\balance
\bibliographystyle{IEEEbib}
\bibliography{strings,refs}

@InProceedings{qin2022finer,
  author    = "Qin, S. and Wang, L. and Li, S. and Lin, Y. and Dang, J.",
  title     = "Finer-grained Modeling Units-based Meta-Learning for Low-resource Tibetan Speech Recognition",
  booktitle = "Proc. Interspeech",
  pages     = "2133-2137",
  year      = "2022",
  doi       = "10.21437/Interspeech.2022-10015"
}

@Article{wu2024mpsa,
  author    = "Wu, C. and Sun, H. and Huang, K. and Wu, L.",
  title     = "MPSA-Conformer-CTC/Attention: A High-Accuracy, Low-Complexity End-to-End Approach for Tibetan Speech Recognition",
  journal   = "Sensors",
  year      = "2024",
  volume    = "24",
  number    = "21",
  pages     = "6824",
  month     = "October",
  doi       = "10.3390/s24216824",
  pmid      = "39517720",
  pmcid     = "PMC11548342"
}

@Article{gong2025tibetan,
  author    = "Gong, Z. and Xu, X. and Zhao, Y.",
  title     = "Tibetan--Chinese speech-to-speech translation based on discrete units",
  journal   = "Scientific Reports",
  year      = "2025",
  volume    = "15",
  pages     = "2592",
  doi       = "10.1038/s41598-025-85782-w"
}

@Misc{liu2023blockwise,
  author       = "Liu, H. and Abbeel, P.",
  title        = "Blockwise Parallel Transformer for Large Context Models",
  howpublished = "arXiv preprint arXiv:2305.19370",
  year         = "2023",
  url          = "https://arxiv.org/abs/2305.19370"
}

@Misc{zeyer2021monotonic,
  author       = "Zeyer, A. and Schl{\"u}ter, R. and Ney, H.",
  title        = "A study of latent monotonic attention variants",
  howpublished = "arXiv preprint arXiv:2103.16710",
  year         = "2021"
}

@InProceedings{radford2023robust,
  author    = "Radford, A. and Kim, J. W. and Xu, T. and Brockman, G. and McLeavey, C. and Sutskever, I.",
  title     = "Robust speech recognition via large-scale weak supervision",
  booktitle = "Proc. ICML",
  pages     = "28492-28518",
  year      = "2023"
}

@InProceedings{moritz2021dual,
  author    = "Moritz, N. and Hori, T. and Roux, J. L.",
  title     = "Dual causal/non-causal self-attention for streaming end-to-end speech recognition",
  booktitle = "Proc. Interspeech",
  pages     = "1822-1826",
  year      = "2021"
}

@InProceedings{xia2025mfla,
  author    = "Xia, Y. and Li, H. and Le, C. and Wang, M. and Sun, Y. and Ma, X. and Qian, Y.",
  title     = "MFLA: Monotonic Finite Look-ahead Attention for Streaming Speech Recognition",
  booktitle = "Proc. Interspeech",
  pages     = "4408-4412",
  year      = "2025",
  doi       = "10.21437/Interspeech.2025-1541"
}

@Misc{gimeno2024comparison,
  author       = "Gimeno-G{\'o}mez, D. and Mart{\'i}nez-Hinarejos, C.-D.",
  title        = "Comparison of Conventional Hybrid and CTC/Attention Decoders for Continuous Visual Speech Recognition",
  howpublished = "arXiv preprint arXiv:2402.13004",
  year         = "2024",
  url          = "https://arxiv.org/abs/2402.13004"
}

@Misc{dighe2023leveraging,
  author       = "Dighe, P. and Su, Y. and Zheng, S. and Liu, Y. and Garg, V. and Niu, X. and Tewfik, A.",
  title        = "Leveraging Large Language Models for Exploiting {ASR} Uncertainty",
  howpublished = "arXiv preprint arXiv:2309.04842",
  year         = "2023",
  url          = "https://arxiv.org/abs/2309.04842"
}

@InProceedings{shivakumar2025rescoring,
  author    = "Shivakumar, P. G. and Kolehmainen, J. and Gourav, A. and Gu, Y. and Gandhe, A. and Rastrow, A. and Bulyko, I.",
  title     = "Speech Recognition Rescoring with Large Speech-Text Foundation Models",
  booktitle = "Proc. ICASSP",
  pages     = "1-5",
  year      = "2025",
  doi       = "10.1109/ICASSP49660.2025.10890616"
}

@Misc{chen2023largescale,
  author       = "Chen, T. and Allauzen, C. and Huang, Y. and Park, D. and Rybach, D. and Huang, W. R. and Cabrera, R. and Audhkhasi, K. and Ramabhadran, B. and Moreno, P. J. and Riley, M.",
  title        = "Large-scale Language Model Rescoring on Long-form Data",
  howpublished = "arXiv preprint arXiv:2306.08133",
  year         = "2023",
  url          = "https://arxiv.org/abs/2306.08133"
}

@InProceedings{udagawa2022effect,
  author    = "Udagawa, T. and Suzuki, M. and Kurata, G. and Itoh, N. and Saon, G.",
  title     = "Effect and Analysis of Large-scale Language Model Rescoring on Competitive ASR Systems",
  booktitle = "Proc. Interspeech",
  pages     = "3919-3923",
  year      = "2022",
  doi       = "10.21437/Interspeech.2022-11123"
}

@InProceedings{zeineldeen2023chunked,
  author    = "Zeineldeen, M. and Zeyer, A. and Schl{\"u}ter, R. and Ney, H.",
  title     = "Chunked Attention-Based Encoder-Decoder Model for Streaming Speech Recognition",
  booktitle = "Proc. ICASSP",
  pages     = "11331-11335",
  year      = "2023",
  url       = "https://api.semanticscholar.org/CorpusID:262012606"
}

@InProceedings{dong2019self,
  author    = "Dong, L. and Wang, F. and Xu, B.",
  title     = "Self-attention aligner: A latency-control end-to-end model for ASR using self-attention network and chunk-hopping",
  booktitle = "Proc. ICASSP",
  year      = "2019"
}

@InProceedings{shi2021emformer,
  author    = "Shi, Y. and Wang, Y. and Wu, C. and Yeh, C. and Chan, J. and Zhang, F. and Le, D. and Seltzer, M.",
  title     = "Emformer: Efficient memory transformer based acoustic model for low latency streaming speech recognition",
  booktitle = "Proc. ICASSP",
  pages     = "6783-6787",
  year      = "2021"
}

@InProceedings{chen2021streaming,
  author    = "Chen, X. and Wu, Y. and Wang, Z. and Liu, S. and Li, J.",
  title     = "Developing real time streaming transformer transducer for speech recognition on large-scale dataset",
  booktitle = "Proc. ICASSP",
  pages     = "5904-5908",
  year      = "2021"
}

@InProceedings{swietojanski2023variable,
  author    = "Swietojanski, P. and Braun, S. and Can, D. and Da Silva, T. F. and Ghoshal, A. and Hori, T. and Hsiao, R. and Mason, H. and McDermott, E. and Silovsky, H. and Travadi, R. and Zhuang, X.",
  title     = "Variable attention masking for configurable transformer transducer speech recognition",
  booktitle = "Proc. ICASSP",
  pages     = "1-5",
  year      = "2023"
}

@InProceedings{gulzar2023ministreamer,
  author    = "Gulzar, H. and Busto, M. R. and Eda, T. and Itoyama, K. and Nakadai, K.",
  title     = "miniStreamer: Enhancing small conformer with chunked context masking for streaming ASR applications on the edge",
  booktitle = "Proc. Interspeech",
  pages     = "3277-3281",
  year      = "2023"
}

@InProceedings{vaswani2017attention,
  author    = "Vaswani, A. and Shazeer, N. and Parmar, N. and Uszkoreit, J. and Jones, L. and Gomez, A. N. and Kaiser, {\L}. and Polosukhin, I.",
  title     = "Attention is All You Need",
  booktitle = "Proc. NeurIPS",
  pages     = "6000-6010",
  year      = "2017"
}

@Misc{brown2020language,
  author       = "Brown, T. B. and Mann, B. and Ryder, N. and Subbiah, M. and Kaplan, J. and Dhariwal, P. and Neelakantan, A. and Shyam, P. and Sastry, G. and Askell, A. and Agarwal, S. and Herbert-Voss, A. and Krueger, G. and Henighan, T. and Child, R. and Ramesh, A. and Ziegler, D. M. and Wu, J. and Winter, C. and Hesse, C. and Chen, M. and Sigler, E. and Litwin, M. and Gray, S. and Chess, B. and Clark, J. and Berner, C. and McCandlish, S. and Radford, A. and Sutskever, I. and Amodei, D.",
  title        = "Language Models are Few-Shot Learners",
  howpublished = "arXiv preprint arXiv:2005.14165",
  year         = "2020",
  url          = "https://arxiv.org/abs/2005.14165"
}

@InProceedings{graves2006ctc,
  author    = "Graves, A. and Fern{\'a}ndez, S. and Gomez, F. and Schmidhuber, J.",
  title     = "Connectionist temporal classification: Labelling unsegmented sequence data with recurrent neural networks",
  booktitle = "Proc. ICML",
  pages     = "369-376",
  address   = "New York, NY, USA",
  year      = "2006"
}

@InProceedings{sheng2025dynamic,
  author    = "Sheng, B. and Yao, J. and Zhang, M. and He, G.",
  title     = "Dynamic Chunking and Selection for Reading Comprehension of Ultra-Long Context in Large Language Models",
  booktitle = "Proc. ACL",
  pages     = "31857-31876",
  year      = "2025",
  doi       = "10.18653/v1/2025.acl-long.1538",
  url       = "https://aclanthology.org/2025.acl-long.1538/"
}

@Article{rumelhart1986backprop,
  author    = "Rumelhart, D. and Hinton, G. and Williams, R.",
  title     = "Learning representations by back-propagating errors",
  journal   = "Nature",
  year      = "1986",
  volume    = "323",
  pages     = "533-536",
  doi       = "10.1038/323533a0"
}

@InProceedings{sharma2025unifying,
  author    = "Sharma, B. and Pandia D, K. S. and Venkatesan, S. and Prakash, J. J. and Kumar, S. and Chetlur, M. and Stolcke, A.",
  title     = "Unifying Streaming and Non-streaming Zipformer-based {ASR}",
  booktitle = "Proc. ACL",
  pages     = "1254-1262",
  year      = "2025",
  doi       = "10.18653/v1/2025.acl-industry.87",
  url       = "https://aclanthology.org/2025.acl-industry.87/"
}

@InProceedings{schultz2013globalphone,
  author    = "Schultz, T. and Vu, N. T. and Schlippe, T.",
  title     = "GlobalPhone: A multilingual text \& speech database in 20 languages",
  booktitle = "Proc. ICASSP",
  pages     = "8126-8130",
  year      = "2013",
  doi       = "10.1109/ICASSP.2013.6639248"
}

@Article{davis1980fbank,
  author    = "Davis, S. B. and Mermelstein, P.",
  title     = "Comparison of parametric representations for monosyllabic word recognition in continuously spoken sentences",
  journal   = "IEEE Transactions on Acoustics, Speech, and Signal Processing",
  year      = "1980",
  volume    = "28",
  number    = "4",
  pages     = "357-366",
  month     = "August",
  doi       = "10.1109/TASSP.1980.1163420"
}

@Misc{graves2012sequencetransduction,
  author       = "Graves, A.",
  title        = "Sequence Transduction with Recurrent Neural Networks",
  howpublished = "arXiv preprint arXiv:1211.3711",
  year         = "2012",
  url          = "https://arxiv.org/abs/1211.3711"
}

@InProceedings{lafferty2001crf,
  author    = "Lafferty, J. D. and McCallum, A. and Pereira, F. C. N.",
  title     = "Conditional Random Fields: Probabilistic Models for Segmenting and Labeling Sequence Data",
  booktitle = "Proc. ICML",
  pages     = "282-289",
  year      = "2001"
}

\end{document}